\documentclass[journal,twoside,web]{IEEEtran}
\usepackage[dvipsnames,svgnames,table,x11names]{xcolor}
\usepackage{cite}
\usepackage{amsmath,amssymb,amsfonts}
\usepackage{algorithmic}
\usepackage{graphicx}
\usepackage{textcomp}
\usepackage{xspace}
\usepackage{booktabs}
\usepackage{multirow}
\usepackage{hyperref}
\usepackage{sidecap}
\usepackage{wrapfig}
\usepackage{subcaption}

\newcommand{\etal}{{\em et al\,.}}       
\newcommand{\eg}{{\em e.g.}}           
\newcommand{\ie}{{\em i.e.}}           

\newcommand{\blue}[1]{#1}
\newcommand{\red}[1]{#1}
\newcommand{\rred}[1]{#1}

\begin{document}

\title{WeakMedSAM: Weakly-Supervised Medical Image Segmentation via SAM with Sub-Class Exploration and Prompt Affinity Mining}
\author{Haoran Wang, Lian Huai, Wenbin Li, Lei Qi, Xingqun Jiang, Yinghuan Shi
\thanks{This work was supported by the NSFC Project (62222604, 62206052), China Postdoctoral Science Foundation (2024M750424), Fundamental Research Funds for the Central Universities (020214380120, 020214380128), State Key Laboratory Fund (ZZKT2024A14), Postdoctoral Fellowship Program of CPSF (GZC20240252), Jiangsu Funding Program for Excellent Postdoctoral Talent (2024ZB242), Young Elite Scientists Sponsorship Program by CAST (2023QNRC001), and Jiangsu Science and Technology Major Project (BG2024031).}
\thanks{Haoran Wang is with the National Key Laboratory for Novel Software Technology and the National Institute of Healthcare Data Science, Nanjing University, Nanjing, Jiangsu, China (e-mail: 502023330059@smail.nju.edu.cn).}
\thanks{Yinghuan Shi is with the National Key Laboratory for Novel Software Technology, National Institute of Healthcare Data Science, Nanjing University, and Nanjing Drum Tower Hospital, Nanjing, Jiangsu, China (e-mail: syh@nju.edu.cn).}
\thanks{Wenbin Li is with the National Key Laboratory for Novel Software Technology, Nanjing University, Nanjing, Jiangsu, China (e-mail: liwenbin@nju.edu.cn).}
\thanks{Lei Qi is with the Pattern Learning and Mining (PALM) Lab, School of Computer Science and Engineering, Southeast University, Nanjing, Jiangsu, China (e-mail: qilei@seu.edu.cn).}
\thanks{Lian Huai and Xingqun Jiang are with BOE Technology Group Co., Ltd., Beijing, China (e-mail: jiangxingqun@boe.com.cn, huailian@boe.com.cn).}
\thanks{The corresponding author is Yinghuan Shi.}
}

\maketitle

\label{sec:abstract}

\begin{abstract}
    We have witnessed remarkable progress in foundation models in vision tasks. Currently, several recent works have utilized the segmenting anything model (SAM) to boost the segmentation performance in medical images, where most of them focus on training an adaptor for fine-tuning a large amount of pixel-wise annotated medical images following a fully supervised manner. 
    In this paper, to reduce the labeling cost, we investigate a novel weakly-supervised SAM-based segmentation model, namely WeakMedSAM. 
    Specifically, our proposed WeakMedSAM contains two modules: 
1) to mitigate severe co-occurrence in medical images, a sub-class exploration module is introduced to learn accurate feature representations.
2) to improve the quality of the class activation maps, our prompt affinity mining module utilizes the prompt capability of SAM to obtain an affinity map for random-walk refinement.
Our method can be applied to any SAM-like backbone, and we conduct experiments with SAMUS and EfficientSAM.
The experimental results on three popularly-used benchmark datasets, \ie, BraTS 2019, AbdomenCT-1K, and MSD Cardiac dataset, show the promising results of our proposed WeakMedSAM.
    Our code is available at \color{magenta}{\url{https://github.com/wanghr64/WeakMedSAM}}.
\end{abstract}

\section{Introduction}

\label{sec:introduction}

The Segment Anything Model (SAM)~\cite{kirillov2023segment} has achieved remarkable success in the field of computer vision, 
with an escalating interest in adopting SAM for various downstream segmentation tasks~\cite{zhang2023survey,zhang2023comprehensive,awais2023foundational}. Among them, the adaptation of SAM to the field of medical image analysis~\cite{huang2024segment}---a longstanding yet important direction---is receiving increasing attention. 

\begin{figure}[t]
  \centering
  \includegraphics[width=\linewidth]{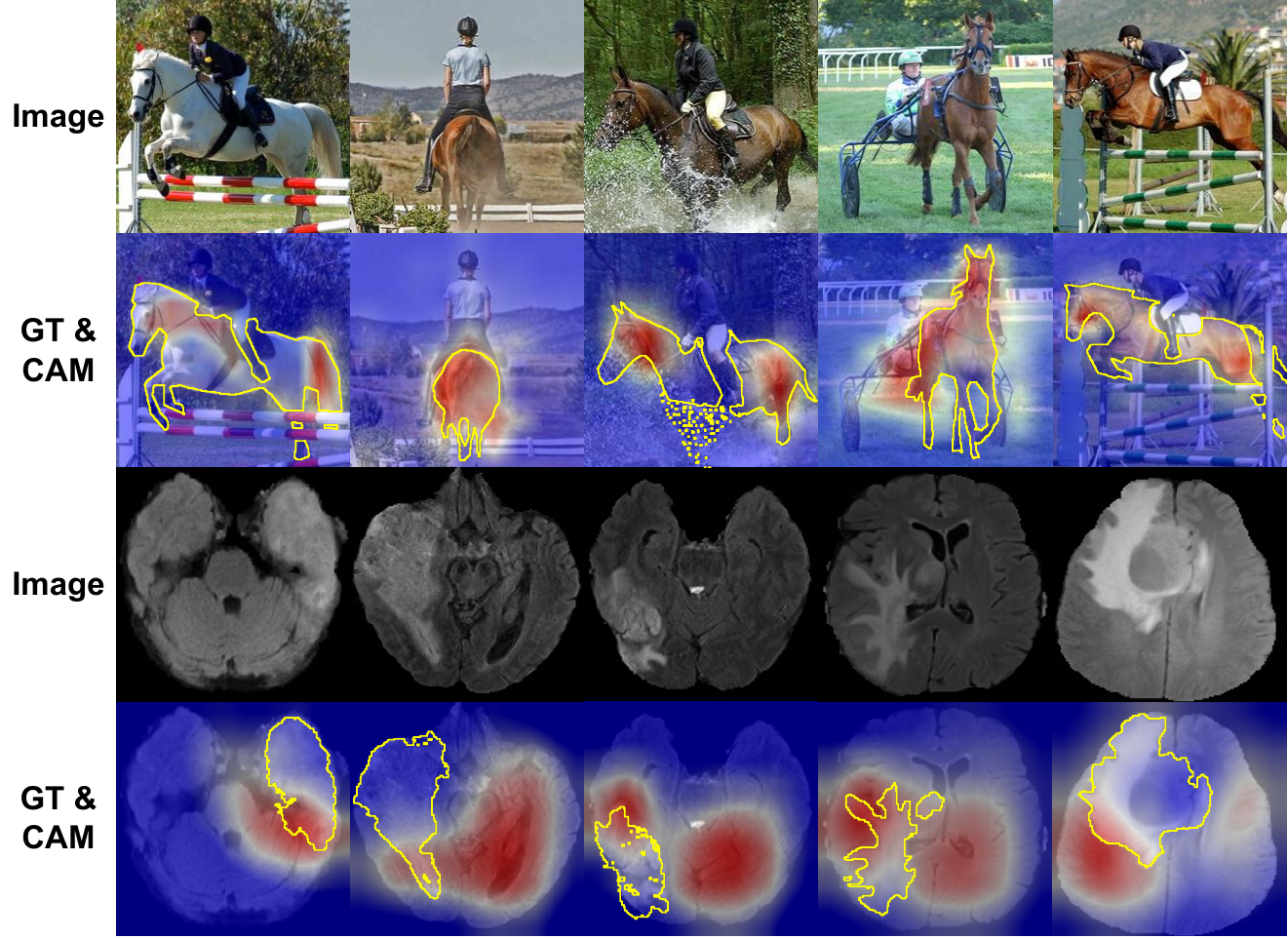}
  \caption{\blue{
  Challenges in weakly-supervised medical image segmentation. 
  The \textcolor[RGB]{220, 220, 0}{\textbf{yellow}} line represents the \textcolor[RGB]{220, 220, 0}{\textbf{ground truth}}.
  Compared to natural images, medical images \blue{suffer} more from co-occurrence phenomena, and CAM tends to activate spurious areas.
   For example, the target of brain tumors incorrectly activates the surrounding edema area, while the target of horses does not mistakenly activate the rider.
  }}
  \label{fig:drawback}
\end{figure}

The common goal of these adapted SAM-based models~\cite{mazurowski2023segment,huang2024segment,zhang2023customized,ma2024segment} is to facilitate effective and efficient adaptation of SAM to solve medical image segmentation. Since the performance of directly utilizing SAM to segment medical images is sometimes unsatisfactory due to the task discrepancy, these methods, \eg, SAMed~\cite{zhang2023customized}, Med-SA~\cite{wu2023medical} and MedSAM~\cite{ma2024segment}, utilizing downstream medical datasets to fine-tune SAM, show the potential to integrate SAM into medical image segmentation. 
\blue{\red{Latest works like Med-SA~\cite{wu2304medical}, One-Prompt SAM~\cite{wu2024one}, and SegAnyPath~\cite{wang2024seganypath} have developed more efficient techniques for adapting SAM to medical images.}}
However, it has been observed that these methods follow fully-supervised paradigm, necessitating extensive medical image datasets with manually delineated pixel-wise segmentation labels. It is known that the acquisition of pixel-wise annotations is a labor-intensive and time-consuming process that requires domain knowledge from experienced physicians, which significantly impedes the deployment of SAM-adapted segmentation in different clinical scenarios. This observation highlights the need for less labor-intensive methods of adapting SAMs to the field of medical imaging. 

As a popular alternative to escape the massive labeling burden, Weakly-Supervised Segmentation (WSS) paradigm~\cite{chan2021comprehensive} is increasingly being recognized. WSS typically employs weak supervision techniques, \eg, image-level labels~\cite{lyu2021weakly,ouyang2020learning,wang2018weakly}, points~\cite{zhao2020weakly,du2023weakly,qu2020weakly}, scribbles~\cite{valvano2021learning,li2024scribformer,wang2018interactive}, or bounding boxes~\cite{rajchl2016deepcut,li2022domain,wang2018interactive}, as a means to circumvent the necessity for comprehensive pixel-wise supervision. Among these weak supervision methods, we in this paper focus on \textit{image-level labels}, due to their simplicity and lack of ambiguity, as image-level labels are unique, whereas other weak supervision methods such as points may introduce possible ambiguity.

Through our empirical and experimental observations, we have identified several challenges associated with SAM-based framework for weakly-supervised segmentation:

\textbf{a)} The segmentation results using weak supervision
primarily rely on inter-class information, \ie the distinctions between different classes (\eg, tumors and healthy tissue, horses and bicycles), which allows each class to have its own task-relevant activation regions. 
Besides, intra-class regions are elements within the same class that often co-occur (\eg, tumors and edema, horses and riders), leading to spurious task-irrelevant activation regions~\cite{chen2022c,ouyang2022causality,xing2021categorical}. 
The impact of co-occurrence is more serious on medical images compared to natural images.  
\blue{\red{For small segmentation targets such as tumors, this phenomenon becomes even more pronounced, and existing WSS methods often struggle to effectively handle such small targets.}}
As depicted in Fig.~\ref{fig:drawback}, the peripheral area of the lesion co-occurs strongly with the target areas, leading to activate inaccurately or even exclusively the erroneous peripheral regions.
This inherent challenge of WSS in medical imaging is also encountered when utilizing models based on SAM.
\textit{Thus, we wonder how to alleviate the mis-activated inter-class regions caused by co-occurrence?}

\textbf{b)} CAMs are primarily used to identify the most discriminative regions within an image, with a possibility to cause either under- or over-segmentation \cite{chan2021comprehensive}. Thus, the refinement of CAMs necessitates the implementation of supplementary strategies. 
However, there seems to be no such thing as a free lunch.
For example, affinity-based methodologies~\cite{ahn2018learning} refine CAM  by training an auxiliary network, thereby introducing additional computational overhead. 
Conversely, methodologies that rely solely on pixels, such as Conditional Random Fields (CRF)~\cite{zheng2015conditional,vemulapalli2016gaussian}, are incapable of incorporating structural information.
\textit{Therefore, is there a method that not only utilizes the existing parameters of SAM, but also leverages medical structural information to refine the CAM?}

To resist the aforementioned challenges, a novel weakly-supervised medical image segmentation framework, namely \textbf{WeakMedSAM}, is proposed in this paper.

Specifically, to address the intra-class co-occurrence in medical images, a feasible way is to divide each primary class (\eg, with tumors and without tumors) into several sub-classes (\eg, latent variations of tumor types) to partition the intra-class representation. Before training, we conduct pre-clustering of sample features that belong to a same primary class to acquire sub-class labels. Subsequently, we combine the sub-class classification task with that of the primary class. The intra-class representations are explicitly learned by the sub-class classification head, so that the primary classification head can achieve an accurate inter-class activation region. 
During our experiments, we were surprised to discover that, optimizing solely the primary class classification head without optimizing the sub-class classification head, could sometimes fortuitously result in concurrent optimizations of sub-class classification loss and CAM quality. This highlights the importance of intra-class representation in enhancing CAM quality and showcases the capability of sub-class mechanism to explicitly acquire and enhance that representation, thereby boosting the precision of class activation.
We refer to this module as \textbf{S}ub-\textbf{C}lass \textbf{E}xploration (\textbf{SCE}).

To maximize the utilization of the parameters and the prompt capability of SAM, affinity maps, which are defined as the relationship between pixels, are acquired for image samples through grid point prompts. Subsequently, the probability matrix derived from the affinity map determines whether the activation region will expand or contract at specific pixel, to procure the final pseudo-labels by executing random walks on CAM.
Our proposed methodology for refining the CAM eliminates the necessity for training an auxiliary network, thereby substantially mitigating the computational expenditure. This module is referred to as \textbf{P}rompt-\textbf{A}ffinity \textbf{M}ining  (\textbf{PAM}).

Our WeakMedSAM, conceived as a plug-and-play module, is applicable to any SAM-based model. To ensure generalizability, we performed experiments using SAMUS~\cite{lin2023samus} and EfficientSAM~\cite{xiong2024efficientsam}. We intentionally refrained from utilizing SAM-based models that have undergone extensive training on large medical datasets like MedSAM~\cite{ma2024segment}, in order to prevent potential biases arising from the inclusion of our experimental datasets in their training data. \blue{Additionally, we demonstrate the robustness of our approach by evaluating it with different pre-trained feature extractors, including models pre-trained on ImageNet and MIMIC-CXR datasets.}

By extensively evaluating on three benchmark datasets---BraTS 2019, AbdomenCT-1K and MSD Cardiac dataset---our method demonstrates promising results. Our method achieves $79.69\%$/$77.25\%$ in Dice and $5.57$/$10.35$ voxels in ASSD utilizing SAMUS and EfficientSAM respectively, compared to newest WSS method with $74.61$ in Dice and $11.91$ voxels in ASSD on BraTS 2019 dataset. \blue{Furthermore, we validate the effectiveness of our method on small target segmentation by modifying BraTS dataset referred to as BraTS-Core, and provide comparisons with fully supervised methods to highlight the practical potential of our approach.}
\begin{figure*}[t]
    \centering
    \includegraphics[width=\textwidth]{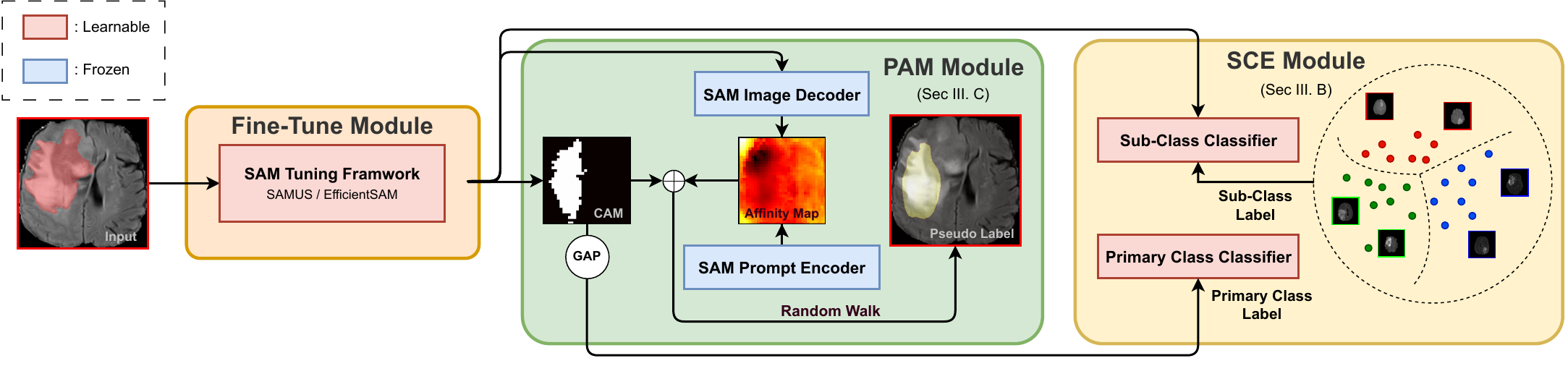}
    \caption{\blue{The overall framework of WeakMedSAM. Before training, WeakMedSAM \red{utilizes} a pretrained network to extract image features and perform pre-clustering. During the training process, all parameters of SAM are frozen. The generated sub-class labels provide additional classification supervision. Then the CAMs are combined with prompt affinity maps for random walks, resulting in the final pseudo-labels.}
    }
    \label{fig:overall}
\end{figure*}

In conclusion, our main contributions are summarized as three folds:
\begin{itemize}
  \item We first attempt to investigate SAM-based weak-supervised medical image segmentation model to simultaneously mitigate the labeling cost and borrow the impressive ability of SAM.
  \item We introduce a sub-class exploration module that effectively alleviates the challenging co-occurrence issue in medical images and consequently yields more precise class activation regions.
  \item We propose a prompt affinity mining module that leverages SAM's existing prompt capability to integrate structural information for refining the CAM without the need for additional training.
\end{itemize}

\section{Related Work}

\label{sec:related-work}

\subsection{Label-Efficient Image Segmentation}

\blue{\red{In the field of label-efficient image segmentation, semi-supervised and few-shot learning approaches have garnered significant attention for their ability to achieve robust segmentation with limited labeled data. Semi-supervised segmentation methods~\cite{qiu2025devil,you2024rethinking,ma2024constructing,wang2024enhancing} leverage both labeled and unlabeled data to enhance model performance. On the other hand, few-shot segmentation~\cite{leng2024self,oliveira2024meta,zhang2024prototype} addresses the challenge of learning from minimal labeled examples by leveraging meta-learning frameworks, which extract class-specific prototypes from support images to guide segmentation on query images.}}

\blue{\red{While semi-supervised and few-shot methods have made significant strides in reducing annotation requirements, our work shifts focus to an even more label-efficient paradigm. Our method utilizes a weakly supervised segmentation framework that relies solely on image-level class labels, further alleviating the burden of data annotation. }}

\subsection{Weakly-Supervised Image Segmentation} Weakly-Supervised Segmentation (WSS) ~\cite{lyu2021weakly,ouyang2020learning,wang2018weakly} is a paradigm that performs image segmentation using only image-level labels. Basically, the initial pseudo labels are usually generated using CAM. However, a common drawback of CAM is its tendency to only activate the most discriminative regions. To overcome this limitation, various training strategies have been proposed in recent studies. 
For instance, some approaches integrate techniques such as erasing~\cite{wei2017object}, online attention accumulation~\cite{jiang2019integral}, and cross-image semantic mining~\cite{sun2020mining} to enhance the segmentation process. Other methods leverage auxiliary tasks to regularize the training objective, including visual word learning~\cite{ru2022weakly} and scale-invariance regularization~\cite{wang2020self}. Additionally, some techniques~\cite{lee2021railroad,yao2021non} utilize additional saliency maps as supervision to suppress background regions and identify non-salient objects effectively. Furthermore, certain methods~\cite{chen2022self,du2022weakly,zhou2022regional} contrast pixel and prototype representations to encourage a more comprehensive activation of object regions.

Unlike these methods, we propose a weakly supervised segmentation approach based on SAM's prompt capability, integrating structural information  in medical images without the need for additional training. \blue{Moreover, our method significantly outperforms the aforementioned WSS approaches in segmenting small targets, such as small tumors, demonstrating superior performance in handling fine-grained structures within medical images.}

\subsection{SAM Tuning for Medical Images} SAM has demonstrated outstanding performance when applied to natural images. However, it encounters challenges in certain medical image segmentation tasks, particularly when dealing with objects characterized by complex shapes, blurred boundaries, small sizes, or low contrast~\cite{mazurowski2023segment,huang2024segment}. To enable SAM to effectively adapt to the medical image domain, various methods~\cite{ma2024segment,zhang2023customized,lin2023samus} have been proposed to fine-tune SAM using downstream medical datasets. Ma \etal~comprised over one million images to develop MedSAM~\cite{ma2024segment}. Also, Zhang \etal~introduced LoRA into SAM for SAMed\cite{zhang2023customized}.
\blue{Alongside these fine-tuning methods above, recently more innovative approaches are now leveraging SAM for medical image segmentation. For example, Medical SAM Adapter (Med-SA)~\cite{wu2304medical} employs a lightweight yet effective adaptation technique to integrate domain-specific medical knowledge into the segmentation model. Moreover, One-Prompt SAM~\cite{wu2024one} skillfully handles unseen tasks during the inference stage with a single prompt, enabling processing in one forward pass. Furthermore, SegAnyPath~\cite{wang2024seganypath} introduces a multi-scale proxy task and an innovative task-guided mixture-of-experts architecture, demonstrating remarkable performance in pathology image segmentation.}
However, it is worth noting that these existing approaches rely on fully supervised labels for fine-tuning SAM, requiring a sufficient number of medical images to be precisely annotated. 

In contrast, our proposed method leverages only image-level class labels, significantly reducing the data annotation costs compared to precise annotation. This aspect is crucial for accurate and easily deployable medical image segmentation in clinical scenarios.

\section{Methodology}

\label{sec:methodology}

As depicted in Fig.~\ref{fig:overall}, our proposed WeakMedSAM consists of two modules: \textbf{a) SCE} (\textbf{S}ub-\textbf{C}lass \textbf{E}xploration module): A weakly supervised fine-tuning of SAM's ViT encoder using image-level labels while exploiting the sub-class features to obtain CAM. \textbf{b) PAM} (\textbf{P}rompt \textbf{A}ffinity \textbf{M}ining): A random walk refinement of the CAM is performed using a prompt affinity map, to enhance the structural representation in medical images. The overall architecture of WeakMedSAM is illustrated in Fig.~\ref{fig:overall} for more details.

\subsection{SAM Fine-tuning Structure}

Our proposed WeakMedSAM approach is compatible with any SAM-based models. In this study, we utilize SAMUS~\cite{lin2023samus} and EfficientSAM~\cite{xiong2024efficientsam}, where SAMUS represents the traditional SAM fine-tuning framework, while EfficientSAM embodies the parameter-efficient SAM fine-tuning framework after knowledge distillation.

\subsubsection{SAMUS} The implementation of SAMUS incorporates a parallel CNN branch into SAM's ViT encoder, utilizing cross-branch attention to enhance the segmentation of medical images. Subsequently, a position adapter and a feature adapter are developed to facilitate the adaptation of SAM from natural to medical domains. The CNN branch consists of a series of interconnected convolutional pooling blocks arranged sequentially. The cross-branch attention module serves as a link between the CNN and ViT branches. All the feature adapters share a consistent structure that includes three primary components: 1) a downward linear projection, 2) an activation function, and 3) an upward linear projection. SAMUS effectively reduces the computational overhead associated with fine-tuning without altering the parameters in SAM itself, thereby representing the traditional fine-tuning framework.
\begin{figure}
 \centering
 \includegraphics[width=\linewidth]{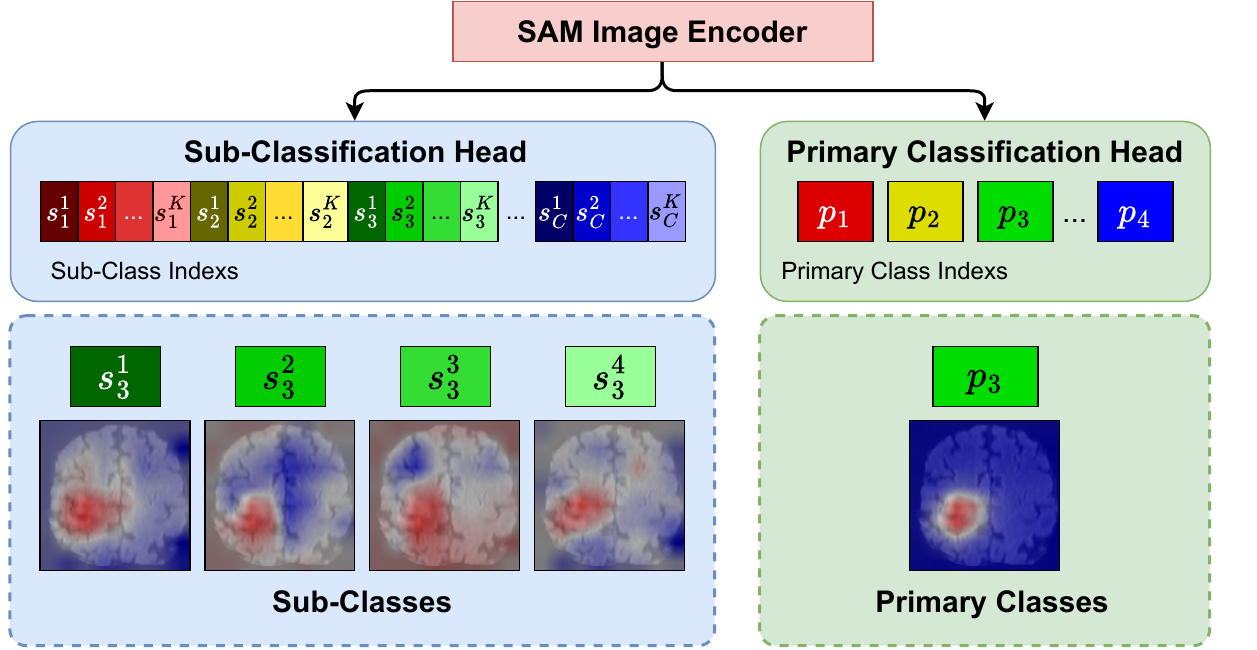}
 \caption{\blue{\red{Activated regions of sub-class and primary class classification. Different sub-classes within the same primary class trigger distinct intra-class discriminative regions. Through the utilization of sub-class classification heads extracting extraneous intra-class information, the primary class classification head is capable of acquiring a more robust inter-class activation representation.}}}
 \label{fig:cluster}
\end{figure}

\subsubsection{EfficientSAM}

EfficientSAM proposes an innovative approach to the pretraining of SAM, known as SAM-leveraged masked image pretraining. This method produces lightweight ViT backbones for segmentation tasks by integrating the well-established MAE~\cite{he2022masked} pretraining technique with the SAM model, thereby facilitating the development of high-quality pretrained ViT encoders. Specifically, EfficientSAM utilizes the SAM encoder to generate feature embeddings and trains a masked image model with lightweight encoders to reconstruct features derived from SAM instead of traditional image patches. EfficientSAM redesigns SAM's ViT encoder to significantly decrease the number of parameters, providing a strategy for the fine-tuning of SAM by knowledge distillation.

\blue{WeakMedSAM is designed to be plug-and-play compatible with any segmentation network based on SAM-like architectures, and its performance and complexity are influenced by the backbone SAM network. To validate the versatility of WeakMedSAM, we tested it on the two SAM-like backbones mentioned above, \red{SAMUS and EfficientSAM}, proving its ability to deliver superior performance across a wide range of SAM-like backbone networks. The subsequent methodological descriptions remain agnostic to the SAM backbone network.}

\subsection{Sub-Class Exploration}

Instead of using pixel-level fully supervised labels, we employ only image-level classification labels to fine-tune SAM. 
To mitigate the co-occurrence phenomenon of WSS in medical images and emphasize inter-class knowledge, we introduce a sub-class classification task supervised by sub-class labels obtained through clustering.
This task explicitly learns the undesired intra-class representations, while allowing the primary classification task activating task-relevant regions.
\blue{\red{This approach is especially significant in medical imaging, where intra-class co-occurrence is prevalent, and it further enhances the model's ability to accurately activate smaller targets like tumors.}}
The SCE module is illustrated in Fig.~\ref{fig:cluster}.

\subsubsection{Obtaining sub-class label} 
\begin{figure}
 \centering
 \includegraphics[width=\linewidth]{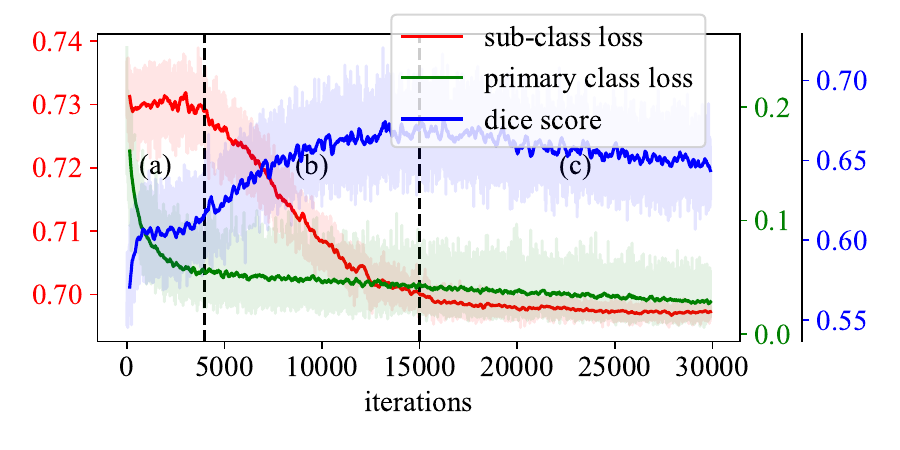}
 \caption{Training process without introducing SCE. Focusing solely on optimizing the primary class classification loss $L_p$, can probably lead to improved sub-class feature representations and more accurate class activation regions.}
 \label{fig:sce_ex}
\end{figure}
For each primary class, denoted as $p_c$, where $c\in\{1,\dots,C\}$, we designate $K$ sub-classes represented as $s_c^k$, where $k\in\{1 ,\dots,K\}$. For every image sample $\boldsymbol x$ with the primary class label $y^c_p \in \{0, 1\}^C$, the corresponding sub-class label for class $c$ is a $0$-$1$ vector of length $K$ represented as $y^c_s\in \{0, 1\}^K$.

Since ground truth labels are not available for sub-classes to optimize directly, we generate pseudo sub-class labels by employing unsupervised clustering. In particular, we conduct clustering for each primary class on the image features extracted from the feature extractor. 
The clustering algorithm can be any unsupervised clustering method such as K-means, and the feature extractor can be any pre-trained image encoder.

\blue{\red{In Section~\ref{sec:ablation}, we further analyze this process, particularly addressing the randomness of the clustering algorithm and the influence of the feature extractor's structure and pre-training dataset, showing that our method exhibits strong robustness to these factors.}}

\subsubsection{Joint training} 

Our ultimate goal is to develop a sub-class classification head $H_s$ parameterized by $\theta_s$, while concurrently sharing the same SAM's image encoder $E$ with the primary classification head $H_p$ parameterized by $\theta_p$. For both $H_p$ and $H_s$, we employ the multi-label binary cross entropy loss as the classification loss $\mathcal L$.

Once we have obtained the pseudo labels $y_s$ for the sub-classes through the aforementioned clustering process, we proceed to jointly optimize the two classifiers $H_p$ and $H_s$:
\begin{equation}
 \min_{\theta_p,\theta_s}\frac{1}{N}\sum^N\mathcal{L}_p\Big(H_p(E(\boldsymbol{x})),y_p\Big)+\lambda\mathcal{L}_s\Big(H_s(E(\boldsymbol{x})),y_s\Big),
\end{equation}
where $E$ represents SAM's image encoder, $N$ represents the total number of images, and $\lambda$ represents a weight used to balance the two loss functions, which we set to $0.5$. 

Through this approach, 
the sub-classification head explicitly explores the intra-class sub-space,
thereby making the primary classification head focused on learning a clean inter-class representation for the purpose of obtaining a more accurate CAM.

\subsubsection{Analysis} 

To demonstrate the effectiveness of the SCE module, we designed the following experiment: without optimizing the parameters of the sub-class classification head, only the SAM image encoder and the primary class classification head are optimized. And we track the primary class classification loss $\mathcal L_p$, sub-class classification loss $\mathcal L_s$, and Dice coefficient of CAM $\mathcal S_{\text{dsc}}$ during the training process. We obtained interesting result that demonstrates the correlation between refined intra-class feature representations and more accurate class activation regions.

As shown in Fig.~\ref{fig:sce_ex}, at the outset of training (part (a) in Fig.~\ref{fig:sce_ex}), $\mathcal L_p$ swiftly converges, yet $\mathcal S_{\text{dsc}}$ reaches a plateau, while $\mathcal L_s$ remains relatively unchanged.
This optimization process is intuitive and aligns naturally with the expected behavior of the model without the SCE module.
\begin{figure}
 \centering
 \includegraphics[width=\linewidth]{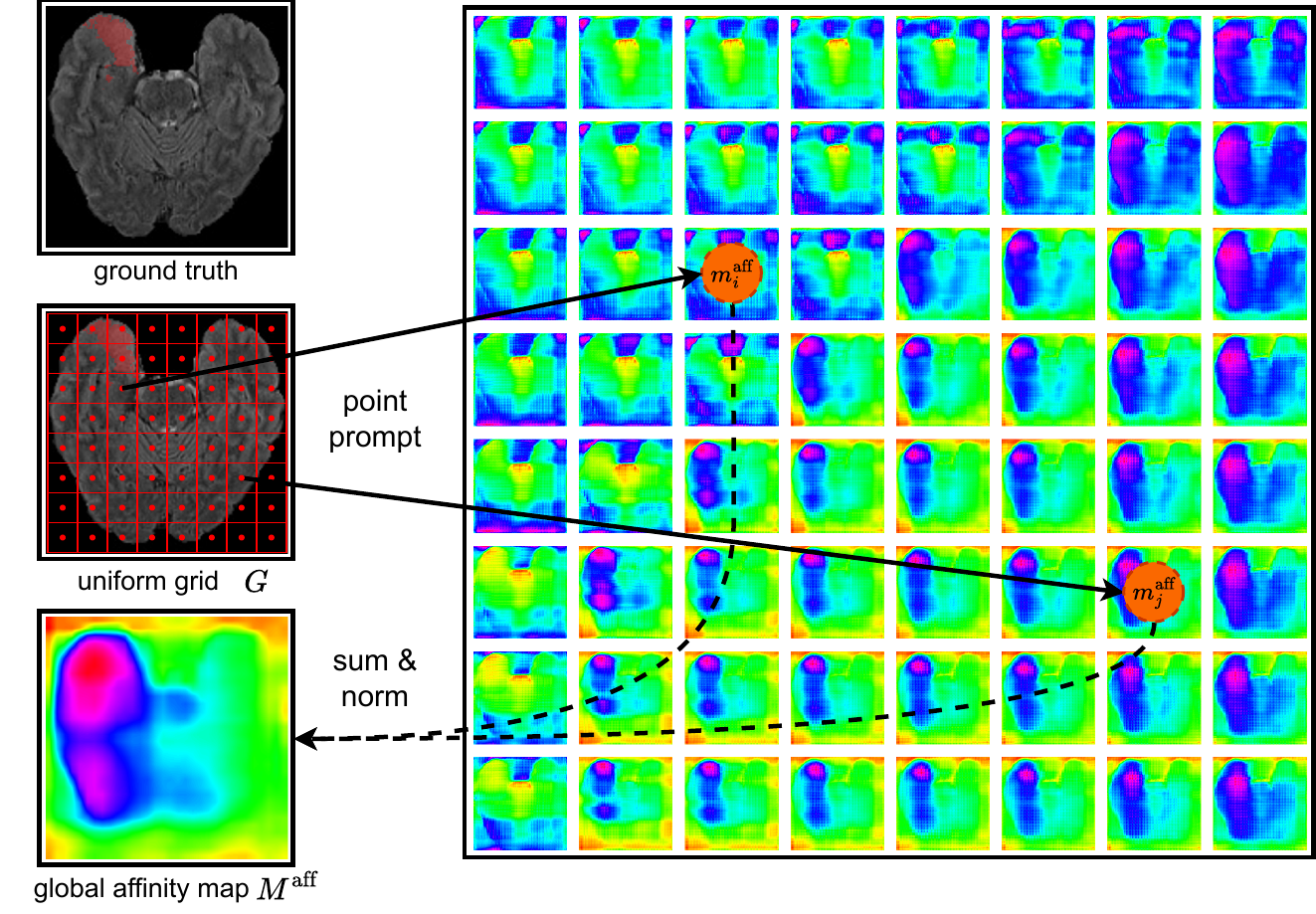}
 \caption{Obtaining affinity map by applying point prompts on SAM.The image is partitioned into a uniform grid whose central points are subjected to the point prompts for the SAM. Then all the prediction maps $m^\text{aff}_i$ are aggregated and normalized to obtain the global affinity map $M^\text{aff}$.}
 \label{fig:affinity}
\end{figure}
Later on (part (b) in Fig.~\ref{fig:sce_ex}), as we persist in solely optimizing $\mathcal L_p$, $\mathcal S_{\text{dsc}}$ experiences significant enhancement, accompanied by a reduction in $\mathcal L_s$. We attribute this to a refined intra-class sub-space feature representation, supported by the observed reduction in the $\mathcal L_s$. This demonstrates that, after mitigating the impact of task-irrelevant intra-class representations on the image encoder, the accuracy of CAM which relies on inter-class information has improved.

Finally (part (c) in Fig.~\ref{fig:sce_ex}), without explicitly learning the intra-class representation, the convergence of $\mathcal L_s$ stops. The ongoing optimization of $\mathcal L_p$ results in overfitting, manifesting in the spurious activation regions, leading to a decrease in $\mathcal S_{\text{dsc}}$.
This further underscores the importance of eliminating intra-class interference to enhance the CAM.

It is crucial to note that not all training processes adhere to this paradigm, and \emph{solely optimizing $L_p$ does not automatically result in further exploration of the intra-class space}. Hence, \emph{the introduction of the SCE module for explicitly optimizing optimize $L_s$}. SCE facilitates the model in acquiring a refined feature representation of the intra-class space, thereby enhancing the accuracy of activation regions.

\subsection{Prompt Affinity Mining}

With the prompt capacity of large models such as SAM, our objective is to achieve class-agnostic affinity between adjacent coordinates on an image without the need for additional training. These affinities are then utilized as transition probabilities in a random walk process, allowing the propagation of CAMs to neighboring regions of the same structural entity. This propagation significantly enhances the quality of CAMs.

\subsubsection{Obtaining affinity map from SAM} 
\begin{figure}
 \centering
 \includegraphics[width=\linewidth]{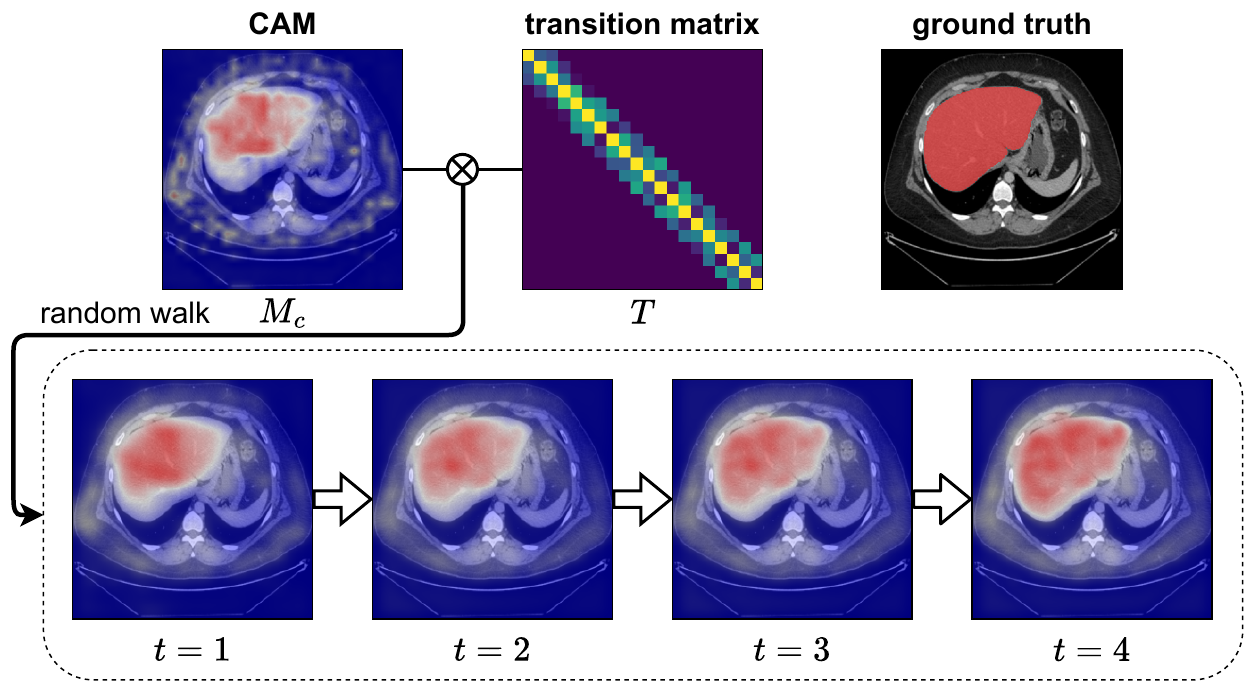}
 \caption{Refining CAM using transition probability matrix $T$ from the affinity map. Utilizing an affinity-based randomized walk approach, the activation area of the CAM becomes more comprehensive.}
 \label{fig:walk}
\end{figure}
The affinity between two coordinates represents the similarity between their class-agnostic structural features, and those in a specific salient region are more meaningful for guiding the refinement. These two properties, \emph{structural similarity} and \emph{spatial locality}, are consistent with SAM's prompt mechanism, in which a point prompt will lead to a \emph{local activation region} with \emph{similar structural features}.

As Fig.~\ref{fig:affinity} depicts, to obtain the affinity map by leveraging the prompt capacity of SAM, initially, the image is partitioned by a uniform grid $G$ with a size of $8\times 8$. Then for each grid region $i$, a \emph{point prompt} is applied at the central point of the grid region, denoted as $p_i$, which yields a mask prediction map $m^{\text{aff}}_i\in\mathbb R^{H\times W}$, the output of the mask decoder. Thus, the overall affinity map of
the image can be defined as follows:
\begin{equation}
 M^{\text{aff}}=\mathtt{norm}\left( \sum_{i\in G}m^{\text{aff}}_i \right),
\end{equation}
where $\mathtt{norm}(\cdot)$ normalizes the affinity values of the graph $m^{\text{aff}}_i$.
Then the global affinity map $M^{\text{aff}}$ will be used to generate the transition probability matrix
$T$ for the proceeding random walk.

\subsubsection{Revising CAMs using the affinity map} The local structural affinities acquired from the SAM are transformed into a transition probability matrix $T$. This matrix facilitates a random walk sensitive to structural regions within the image, promoting the dispersion of activation scores within these regions, as shown in Fig.~\ref{fig:walk}.

The structural affinity between a pair of feature vectors is characterized by their $L_1$ distance. Specifically, we denote the structural affinity between features $i$ and $j$ as $a_{ij}$, which is calculated as follows:
\begin{equation}
 a_{ij}=\exp\left( -\lVert M^{\text{aff}}_i-M^{\text{aff}}_j \rVert_1 \right).
\end{equation}

Note that the affinities are only computed between features within a local circle of radius $\gamma$. Here we take $\gamma=5$. The calculated affinities collectively form an affinity matrix $A$, where the diagonal elements are set to $1$. From this affinity matrix, the transition probability matrix $T$ for the random walk process is derived using the following procedure:
\begin{equation}
 T=D^{-1}A^{\circ\beta},~~\text{where}~D_{ii}=\sum_j A_{ij}^{\circ\beta},
 \label{eq:t_matrix}
\end{equation}
where the hyper-parameter $\beta$ is assigned a value greater than $1$. By raising the original affinity matrix $A$ to the power of $\beta$, denoted as $A^{\circ\beta}$, we effectively suppress insignificant affinities within $A$. This adjustment enables a more cautious propagation of the random walk process. Additionally, the diagonal matrix $D$ is calculated to facilitate row-wise normalization of $A^{\circ\beta}$.

By employing the transition probability matrix $T$, the process of the affinity propagation is achieved through a series of random walk operations. Specifically, the CAMs are multiplied by $T$ to perform this propagation. This iterative propagation procedure continues until the predefined number of iterations $t$ is attained. As a result, the revised CAM of class $c$, denoted as $M^*_c$, is obtained using the following expression:
\begin{equation}
 \text{vec}(M^*_c)=T^{\circ t}\cdot \text{vec}(M_c),~~\forall c\in \{1,\dots,C\},
 \label{eq:random_walk}
\end{equation}
where the operation $\text{vec}(\cdot)$ represents the vectorization of a matrix, and $t$ denotes the number of iterations.

\begin{table*}[t]
 \caption{The comparison with other baseline methods on BraTS 2019, AbdomenCT-1K and MSD Cardiac dataset. $\uparrow$ indicates that the higher the better. $\downarrow$ indicates that the lower the better.
 }
 \label{tab:comparison}
\centering
 \begin{tabular}{c|c|cccc}
 \toprule
 Dateset & Methods & DSC (\%)$\uparrow$ & Jaccard (\%)$\uparrow$ & ASSD (voxel)$\downarrow$ & HD95 (voxel)$\downarrow$ \\ \midrule
 \multirow{8}{*}{BraTS 2019} & CAM~\cite{zhou2016learning} & $63.42$ & $57.97$ & $25.85$ & $39.60$ \\
 & PSA~\cite{ahn2018learning} & $73.02$ & $67.69$ & $19.91$ & $35.97$ \\
 & SEAM~\cite{wang2020self} & $73.70$ & $67.94$ & $18.33$ & $39.63$ \\
 & AFA~\cite{ru2022learning} & $72.02$ & $64.22$ & $12.45$ & $33.14$ \\
 & MCT~\cite{xu2022multi} & $66.62$ & $59.35$ & $17.95$ & $37.07$ \\
 & SIPE~\cite{chen2022self} & $74.27$ & $69.91$ & $14.05$ & $35.80$ \\
 & TOCO~\cite{ru2023token} & $74.61$ & $69.13$ & $11.91$ & $32.69$ \\
 & \cellcolor[HTML]{EFEFEF}\textbf{Ours (EfficientSAM)} &\cellcolor[HTML]{EFEFEF}$\boldsymbol{77.25}$ &\cellcolor[HTML]{EFEFEF}$\boldsymbol{72.58}$ & \cellcolor[HTML]{EFEFEF}$\boldsymbol{10.35}$ & \cellcolor[HTML]{EFEFEF}$\boldsymbol{30.86}$ \\ 
 & \cellcolor[HTML]{EFEFEF}\textbf{Ours (SAMUS)} & \cellcolor[HTML]{EFEFEF}$\boldsymbol{79.69}$ & \cellcolor[HTML]{EFEFEF}$\boldsymbol{74.06}$ & \cellcolor[HTML]{EFEFEF} $\boldsymbol{5.57}$ & \cellcolor[HTML]{EFEFEF}$\boldsymbol{28.34}$ \\ 
 \midrule
 \multirow{8}{*}{AbdomenCT-1K} & CAM~\cite{zhou2016learning} & $58.38$ & $53.30$ & $28.59$ & $91.15$ \\
 & PSA~\cite{ahn2018learning} & $66.08 $ & $62.97$ & $21.49$ & $80.11$ \\
 & SEAM~\cite{wang2020self} & $65.31$ & $60.14$ & $22.77$ & $72.61$ \\
 & AFA~\cite{ru2022learning} & $70.87$ & $65.52$ & $21.59$ & $75.72$ \\
 & MCT~\cite{xu2022multi} & $66.10$ & $61.60$ & $27.64$ & $82.25$ \\
 & SIPE~\cite{chen2022self} & $67.20$ & $62.71$ & $19.38$ & $70.78$ \\
 & TOCO~\cite{ru2023token} & $65.82$ & $60.89$ & $22.25$ & $75.13$ \\
 & \cellcolor[HTML]{EFEFEF}\textbf{Ours (EfficientSAM)} & \cellcolor[HTML]{EFEFEF}$\boldsymbol{71.15}$ & \cellcolor[HTML]{EFEFEF}$\boldsymbol{66.73}$ & \cellcolor[HTML]{EFEFEF}$\boldsymbol{18.62}$ & \cellcolor[HTML]{EFEFEF}$\boldsymbol{69.57}$ \\ 
 & \cellcolor[HTML]{EFEFEF}\textbf{Ours (SAMUS)} & \cellcolor[HTML]{EFEFEF}$\boldsymbol{75.87}$ & \cellcolor[HTML]{EFEFEF}$\boldsymbol{72.17}$ & \cellcolor[HTML]{EFEFEF}$\boldsymbol{16.35}$ & \cellcolor[HTML]{EFEFEF}$\boldsymbol{62.75}$ \\ 
 \midrule
 \multirow{8}{*}{MSD Cardiac} & CAM~\cite{zhou2016learning} & $50.04$ & $42.58$ & $32.58$ & $105.54$ \\
 & PSA~\cite{ahn2018learning} & $53.33$ & $46.15$ & $28.40$ & $86.01$ \\
 & SEAM~\cite{wang2020self} & $55.47$ & $48.80$ & $24.48$ & $77.25$ \\
 & AFA~\cite{ru2022learning} & $53.34$ & $47.91$ & $26.42$ & $79.28$ \\
 & MCT~\cite{xu2022multi} & $52.98$ & $45.93$ & $27.25$ & $85.52$ \\
 & SIPE~\cite{chen2022self} & $56.59$ & $50.02$ & $24.29$ & $69.90$ \\
 & TOCO~\cite{ru2023token} & $54.67$ & $48.21$ & $25.04$ & $76.97$ \\
 & \cellcolor[HTML]{EFEFEF}\textbf{Ours (EfficientSAM)} & \cellcolor[HTML]{EFEFEF}$\boldsymbol{57.62}$ & \cellcolor[HTML]{EFEFEF}$\boldsymbol{51.21}$ & \cellcolor[HTML]{EFEFEF}$\boldsymbol{24.09}$ & \cellcolor[HTML]{EFEFEF}$\boldsymbol{68.94}$ \\ 
 & \cellcolor[HTML]{EFEFEF}\textbf{Ours (SAMUS)} & \cellcolor[HTML]{EFEFEF}$\boldsymbol{58.05}$ & \cellcolor[HTML]{EFEFEF}$\boldsymbol{52.67}$ & \cellcolor[HTML]{EFEFEF}$\boldsymbol{22.99}$ & \cellcolor[HTML]{EFEFEF}$\boldsymbol{63.99}$ \\
 \bottomrule
\end{tabular}
\end{table*}

\subsection{Complexity Discussions}

WeakMedSAM is an extension of any SAM-based model. In contrast to the original SAM model, WeakMedSAM integrates two linear layers specifically for the Sub-Class Exploration (SCE) module---the primary class classification header and the subclass classification header, adding minimal computational overhead. Notably, the Prompt Affinity Mining (PAM) module any additional model parameters. The computational overhead of this module only includes the prompt encoder and the mask decoder, which are lightweight and can be even executed in a web browser~\cite{kirillov2023segment}. This design ensures that our approach is well-suited for both training and inference processes, making it feasible to execute on various GPU configurations without requiring extensive computational resources. \blue{In Section~\ref{sec:numeral:complexity}, we provide a further numerical discussion on the complexity of the proposed method, detailing its efficiency and scalability across different computational environments.}

\section{Experiments}

\label{sec:experiments}

\subsection{Experimental Settings}

\begin{figure*}[tb]
 \centering
 \includegraphics[width=\linewidth]{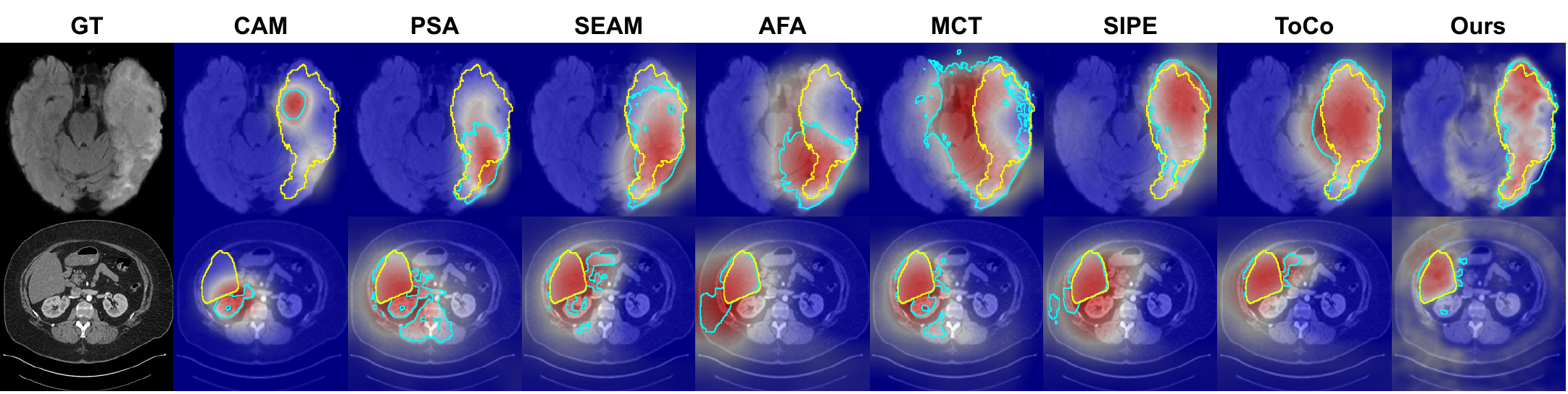}
 \caption{\blue{Visualization of WeakMedSAM and other WSS methods on BraTS 2019 dataset and AbdomenCT-1K dataset.
 The \textcolor[RGB]{220, 220, 0}{\textbf{yellow}} line represents the \textcolor[RGB]{220, 220, 0}{\textbf{ground truth}}, and the \textcolor[RGB]{0, 200, 200}{\textbf{cyan}} line indicates the \textcolor[RGB]{0, 200, 200}{\textbf{segmentation result}}.}
 }
 \label{fig:all-vis-2}
\end{figure*}

\subsubsection{Datasets} We conducted our research by utilizing three widely used datasets: BraTS 2019~\cite{menze2014multimodal}, AbdomenCT-1K~\cite{ma2021abdomenct}, and MSD Cardiac dataset~\cite{antonelli2022medical}.

The BraTS 2019 dataset consists of a total of 335 multi-modality scans, each accompanied by expert segmentation masks. These scans encompass four modalities: T1, T1c, T2, and FLAIR, while here we only utilize the FLAIR modality. The dataset was specifically utilized for a binary segmentation task, where our focus was on distinguishing between healthy and unhealthy targets.
The AbdomenCT-1K dataset is a comprehensive and diverse abdominal CT organ segmentation dataset comprising over 1000 CT scans sourced from 12 medical centers, encompassing multi-phase, multi-vendor, and multi-disease cases. Our goal is four of these abdominal organs: liver, kidney, spleen, pancreas.
The MSD Cardiac dataset consists of MRI images specifically designed for the task of left ventricle segmentation. It contains a total of 30 patient cases, each consisting of multiple 2D MRI slices.

We randomly partitioned it into three subsets: training, validation, and testing sets, in an 8:1:1 ratio on a patient-by-patient basis. In our analysis, we treated individual slices of the 3D-MRI scans as 2D images, focusing on the segmentation task at the slice level.

\blue{To further validate the segmentation performance of our method on small targets like small tumors, we discarded the peritumoral edema in the BraTS dataset and retained only the tumor regions. We refer to this dataset as BraTS-Core, which follows the same partitioning scheme and label definitions as the vanilla BraTS dataset. Detailed descriptions of BraTS-Core dataset are provided in Section~\ref{sec:small-target}.}

\subsubsection{Network architectures} We maintain the prompt encoder and mask decoder of SAM in a frozen state, thereby exclusively training the SAM's image encoder by employing both SAMUS and EfficientSAM. 
For EfficientSAM, we use the ViT-Ti encoder as the backbone with the minimum number of parameters.
Notably, our proposed WeakMedSAM can be used plug-and-play for any SAM-based model.
Without compromising generality, the subsequent analytical experiments are conducted within SAMUS.

Instead of utilizing the output from the encoder's final neck layer, we opt to use the output derived from the transformer block as the image embedding for classification, achieving a better representation for class activation. Furthermore, both the primary and sub-class classification heads employ a $1\times1$ convolutional layer. For the segmentation network trained on the pseudo labels obtained from WeakMedSAM, we use the U-Net~\cite{ronneberger2015u} network, which is widely used in medical image segmentation tasks.

\subsubsection{Implementation details} We conducted our work using Python and the PyTorch framework. The implementation involved running the codes on four NVIDIA GTX 2080Ti GPUs. The learning rate followed a one-cycle policy up to $10^{-4}$ and then decayed for subsequent iterations. The batch size was set to 24, and the total number of epochs was 10. The images were cropped to a size of 256. Additionally, we used a constant weight of $\lambda=0.5$ for the sub-class loss.

\subsubsection{Evaluation metrics} To assess the quality of these pseudo labels, we employ four evaluation metrics: the Dice coefficient (DSC), the Jaccard index, average symmetric surface distance (ASSD) and Hausdorff distance 95\% percentile (HD95). The DSC and the Jaccard index are widely used pixel-level similarity measures that provide insights into the performance of segmentation models. On the other hand, the ASSD and HD95 metric quantify the accuracy of segmentation results by measuring the surface distance between the predicted results and the ground truth labels.

\subsection{Comparison with Other Methods}

\subsubsection{Comparison with WSS methods} 
In order to evaluate the effectiveness of our proposed method WeakMedSAM, we applied a threshold to the CAMs to obtain pseudo-labels, which were then used to train the segmentation network and obtain the final segmentation results. We compared it with recent weakly supervised segmentation methods, as shown in Tab.~\ref{tab:comparison} and Fig.~\ref{fig:all-vis-2}. Our method shows excellent accuracy across a range of datasets and demonstrates outstanding performance on two distinct SAM-based networks, consequently validating the generality of our approach.

\begin{figure*}[t]
  \centering
  \includegraphics[width=\linewidth]{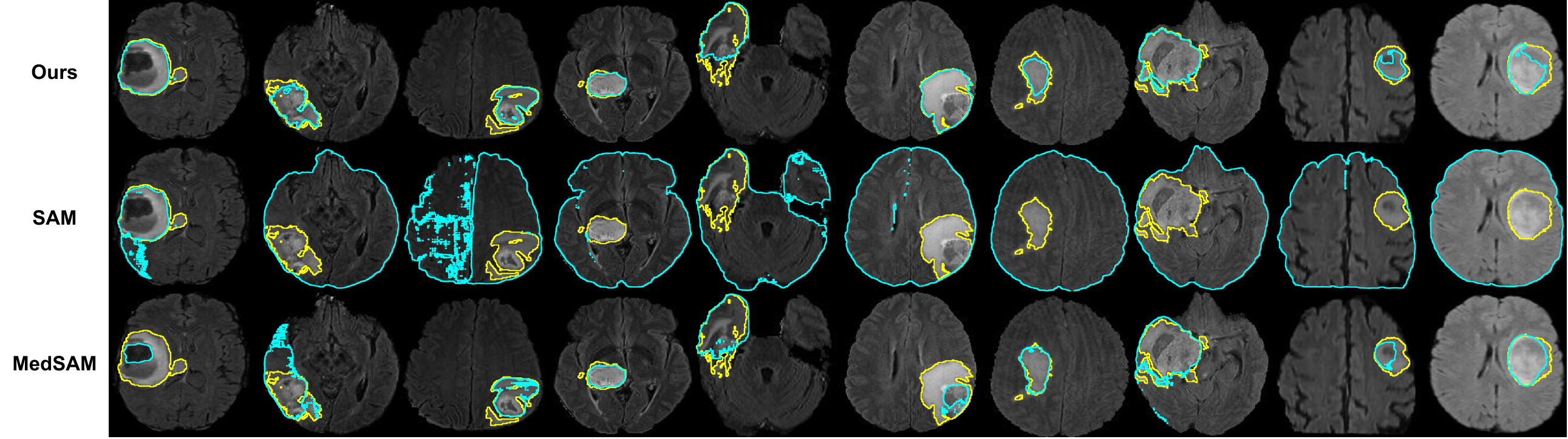}
  \caption{\blue{Comparison between SAM, MedSAM and WeakMedSAM. 
  The \textcolor[RGB]{220, 220, 0}{\textbf{yellow}} line represents the \textcolor[RGB]{220, 220, 0}{\textbf{ground truth}}, and the \textcolor[RGB]{0, 200, 200}{\textbf{cyan}} line indicates the \textcolor[RGB]{0, 200, 200}{\textbf{segmentation result}}.
  SAM in most cases tends to result in severe over-segmentation or under-segmentation due to the task discrepancy.
  MedSAM while showing relatively good performance, is sensitive to the quality of prompts.
  }}
  \label{fig:vanilla}
 \end{figure*}

\subsubsection{Comparison with interactive SAM-like methods}

\begin{table}[t]
  \caption{{Comparison with SAM-like interactive segmentation methods using central point prompt on BraTS 2019 dataset.}}
  \label{tab:interactive}
  \centering
  \begin{tabular}{c|cccc}
  \toprule
  Method & \blue{DSC $\uparrow$} & \blue{Jaccard $\uparrow$} & \blue{ASSD $\downarrow$} & \blue{HD95 $\downarrow$}\\ \midrule
  SAM & $55.68$ & $49.33$ & $32.91$ & $76.29$ \\
  MedSAM & $79.12$ & $73.98$ & $6.33$ & $33.19$ \\
  \cellcolor[HTML]{EFEFEF}Ours & \cellcolor[HTML]{EFEFEF}$\boldsymbol{79.69}$ & \cellcolor[HTML]{EFEFEF}$\boldsymbol{74.06}$ & \cellcolor[HTML]{EFEFEF}$\boldsymbol{5.57}$ & \cellcolor[HTML]{EFEFEF}$\boldsymbol{28.34}$ \\ \bottomrule
  \end{tabular}
 \end{table}

While SAM isn't specifically trained on medical image datasets, it demonstrates potential for zero-shot generalization~\cite{wald2023sam}. Particularly, SAM's interactive segmentation, guided by prompts such as points and bounding boxes, can greatly enhance the end-user experience. Leveraging a visual foundation model for interactive segmentation holds promising potential in reducing heavy pixel-level labeling costs.

Therefore, we propose a hypothetical scenario of interactive segmentation where an expert employs visual prompts to specify regions of interest. To equate the cost of this interactive segmentation with our image-level label acquisition costs, we utilize a single point prompt. This coordinates of this prompt correspond to the central point of the segmentation label used in SAM's interactive segmentation.

In recent studies, many efforts have focused on fine-tuning SAM on large-scale medical image datasets. Employing these fine-tuned models for interactive segmentation is likely to be of greater significance in actual clinical applications. Consequently, we have also included MedSAM in our comparisons with SAM.

The findings reveal that SAM exhibits robust zero-shot generalization in some samples, as shown in Fig.~\ref{fig:vanilla}. 
However, in most instances, due to the absence of task-relevant information, the segmentation results of SAM primarily rely solely on pixel information, which is severely lacking in medical images. Thus SAM tends to interpret blank regions as background rather than non-target areas, leading to severe over-segmentation or under-segmentation. Therefore, the direct application of SAM for medical image segmentation is not optimal. In contrast, MedSAM shows substantial improvement in overall segmentation performance compared to SAM. While it performs slightly weaker than our WeakMedSAM, it can achieve strong results with the addition of a minimal number of extra interactive prompts. 

Importantly, the use of class-level labels within a weakly supervised framework eliminates the need for interactive involvement from clinicians during the inference phase. The entire inference process of WeakMedSAM is fully automated, further reducing the time required for segmentation tasks.

\subsection{Ablation Studies and Analysis}
\label{sec:ablation}

\begin{table}[t]
  \caption{{The ablation study for each proposed module of WeakMedSAM on BraTS 2019 dataset.} \checkmark indicates that the corresponding module is used.}
  \label{tab:ablation}
  \centering
  \begin{tabular}{cc|cccc}
  \toprule
  SCE & PAM & \blue{DSC $\uparrow$} & \blue{Jaccard $\uparrow$} & \blue{ASSD $\downarrow$} & \blue{HD95 $\downarrow$} \\ \midrule
  & & $68.13$ & $53.98$ & $12.12$ & $38.33$ \\
  \checkmark & & $74.91$ & $69.82$ & $6.31$ & $32.09$ \\
  & \checkmark & $76.31$ & $71.03$ & $5.63$ & $30.81$ \\
  \cellcolor[HTML]{EFEFEF}\checkmark & \cellcolor[HTML]{EFEFEF}\checkmark & \cellcolor[HTML]{EFEFEF}$\boldsymbol{79.69}$ & \cellcolor[HTML]{EFEFEF}$\boldsymbol{74.06}$ & \cellcolor[HTML]{EFEFEF}$\boldsymbol{5.57}$ & \cellcolor[HTML]{EFEFEF}$\boldsymbol{28.34}$ \\ \bottomrule
  \end{tabular}
  
  \end{table}

The quantitative outcomes of the ablation analysis are presented in Tab.~\ref{tab:ablation}. This table illustrates that our backbone, which is based on ViT-b SAM, secures a DSC of $68.13\%$ on the BraTS 2019 dataset. The introduction of the proposed Sub-Class Classification and Prompt Affinity Mining modules considerably elevates the DSC to $74.91\%$ and $76.31\%$ respectively. When these two models are combined, the performance of the model escalates to $79.69\%$. 
\begin{table}[t]
  \caption{Different backbone of feature extraction on BraTS 2019 dataset.}
  \label{tab:cluster-backbone}
  \centering
  \begin{tabular}{c|cccc}
  \toprule
  Backbone & \blue{DSC $\uparrow$} & \blue{Jaccard $\uparrow$} & \blue{ASSD $\downarrow$} & \blue{HD95 $\downarrow$} \\ \midrule
  \cellcolor[HTML]{EFEFEF}ResNet18 & \cellcolor[HTML]{EFEFEF}$\boldsymbol{79.69}$ & \cellcolor[HTML]{EFEFEF}$\boldsymbol{74.06}$ & \cellcolor[HTML]{EFEFEF}$\boldsymbol{5.57}$ & \cellcolor[HTML]{EFEFEF}$\boldsymbol{28.34}$ \\
  ResNet50 & $79.72$ & $74.21$ & $5.34$ & $27.79$ \\
  ViT-b-16 & $79.81$ & $74.44$ & $5.41$ & $26.65$ \\
  ViT-h-14 & $79.00$ & $73.29$ & $6.03$ & $32.16$ \\ \bottomrule
  \end{tabular}
 \end{table}
\subsubsection{Different backbones for SCE's feature extraction}
\label{sec:backbone-sce}
Due to the unavailability of ground truth labels for sub-classes, we create pseudo labels using unsupervised clustering.
Specifically, clustering is performed for each primary class based on the image features extracted from the feature extractor.

A potential issue is that the performance of the backbone network performing feature extraction may ultimately affect the final performance of sub-class classification.
We performed pre-clustering with different pre-trained networks. 

As shown in Tab.~\ref{tab:cluster-backbone}, our network is not particularly sensitive to the performance of the backbone network.
Consequently, to reduce computational overhead, we employ ResNet18 as the feature extractor during the pre-clustering phase.

\subsubsection{Different pre-training datasets for SCE's feature extractor}

\blue{
While our experiments utilized ImageNet pre-trained models for the SCE module, we acknowledge the importance of considering medical image pre-trained feature extractor. However, identifying or training a model for various medical image modalities remains a challenge.}

\begin{figure}[t]
    \centering
    \captionsetup[subfigure]{labelfont={color=black}}
    \subfloat[\blue{ImageNet}]{\includegraphics[width=0.45\linewidth]{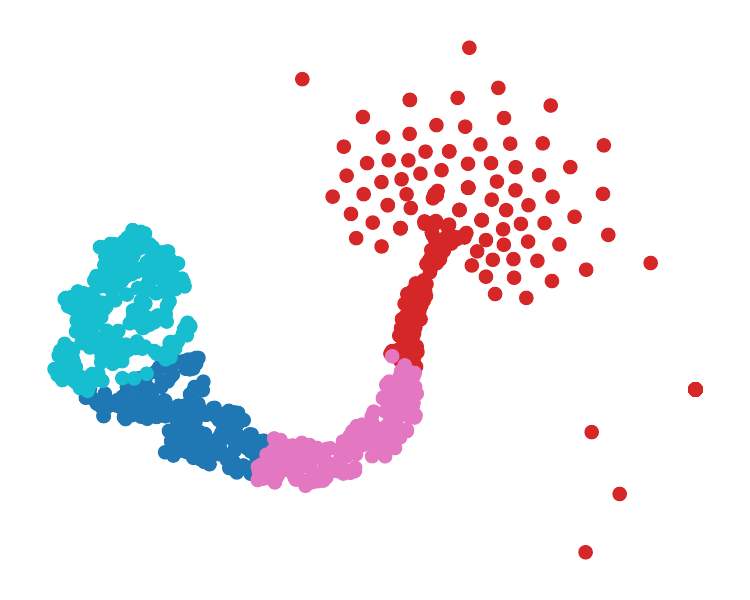}}
    \subfloat[\blue{MIMIC-CXR}]{\includegraphics[width=0.45\linewidth]{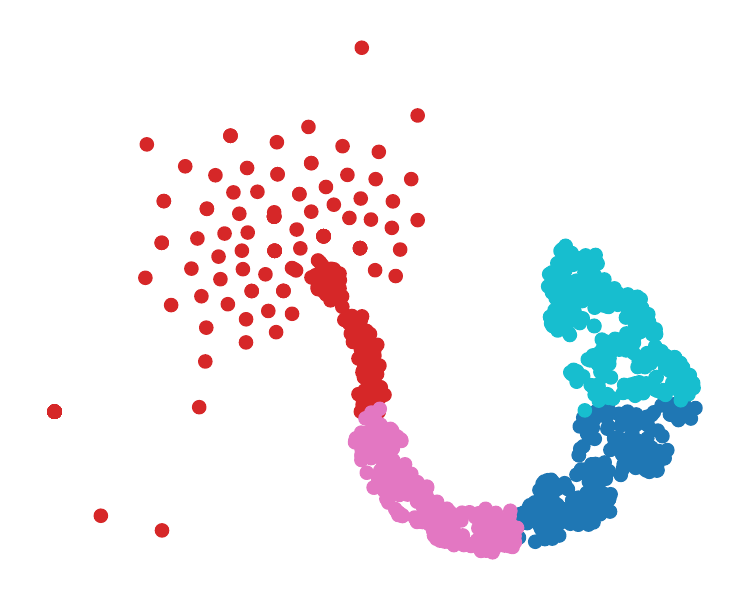}}
    \captionsetup[subfigure]{labelfont={color=black}}
    \caption{\blue{Visualization of image features after SCE clustering on BraTS dataset. \rred{Each color corresponds to a cluster group.} Both the feature extractors, pre-trained on ImageNet (left) and MIMIC-CXR (right), produce \rred{well-separated and} meaningful feature representations for cross-modal data like BraTS.}}
    \label{fig:backbone-trained-on}
\end{figure}

\begin{table}[t]
    \centering
    \caption{\blue{Segmentation performance of the SCE module using feature extractors pre-trained on various datasets.}}
    \label{tab:backbone-trained-on}
    \blue{
    \begin{tabular}{c|cccc}\toprule
          Pre-train Dataset        & DSC $\uparrow$ & Jaccard $\uparrow$ & ASSD $\downarrow$ & HD95 $\downarrow$ \\ \midrule
      \cellcolor[HTML]{EFEFEF} ImageNet   & \cellcolor[HTML]{EFEFEF}\textbf{79.78} & \cellcolor[HTML]{EFEFEF}\textbf{73.74} & \cellcolor[HTML]{EFEFEF}\textbf{7.04} & \cellcolor[HTML]{EFEFEF}\textbf{28.91} \\
       MIMIC-CXR  & 78.14 & 73.17 & 8.97 & 29.46 \\ \bottomrule
    \end{tabular}}
\end{table}

\blue{Existing literature demonstrates the effectiveness of ImageNet pre-trained feature extractors in downstream tasks even with significant domain shifts~\cite{marmanis2015deep,muhammad2018pre,boudiaf2022underwater,szymak2020effectiveness} as well as in medical imaging~\cite{xie2018pre,morid2021scoping,liu2024swin}. Given that our SCE module is not sensitive to feature extractor capacity demonstrated in Section~\ref{sec:backbone-sce}, employing ImageNet pre-trained feature extractors is a justifiable and practical approach.}

\blue{To further analysis this, we evaluated SCE using ViT models pre-trained on both ImageNet and the MIMIC-CXR~\cite{johnson2019mimic} dataset provided by Medical-MAE~\cite{xiao2023delving}. Fig.~\ref{fig:backbone-trained-on} presents the clustering results of SCE using ImageNet and MIMIC-CXR pre-trained ViT on BraTS dataset. And Tab.~\ref{tab:backbone-trained-on} shows the segmentation performance using these extractors. These results demonstrate the robustness of SCE to the pre-training dataset of the feature extractor.}

\begin{figure}
  \centering
  \includegraphics[width=0.65\linewidth]{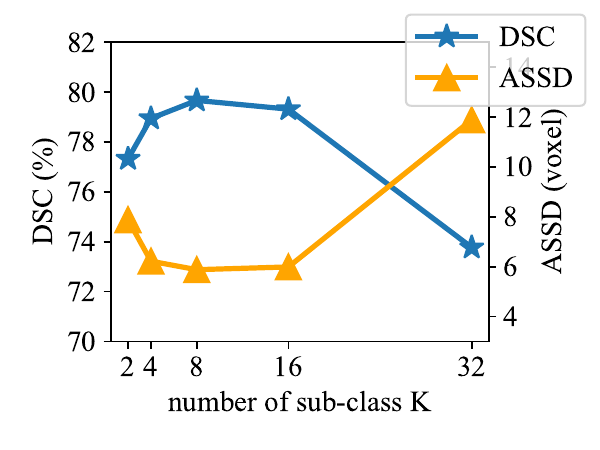}
  \caption{\blue{Comparison between different numbers of sub-classes $K$.
  As the number of sub-classes increases, there is a corresponding enhancement in the performance of the model. Nevertheless, an excessively high number of sub-classes is found to negatively affect the model's performance.}
  }
  \label{fig:number-sub-classes}
\end{figure}

\subsubsection{Different number of sub-classes}

The intention behind introducing the sub-class classification task is to explicitly define the intra-class information, which could potentially distort the activation regions.
This is achieved by learning through the sub-class classification head, thereby enabling the primary class classification head to focus solely on acquiring of inter-class information.

Nevertheless, the number of sub-classes associated with each primary class is manually determined during the pre-clustering process.
This implies that varying numbers of sub-classes may influence the model's performance.
Through experimentation with different numbers of sub-classes, as Fig.~\ref{fig:number-sub-classes} shows, we observed that optimal performance was achieved with $8$ sub-classes.

As a prospective area of study, it would be beneficial to devise an adaptive methodology for determining the number of sub-classes. This could minimize the occurrence of superfluous sub-classes, thereby enhancing the efficiency of the approach.

\subsubsection{Analysis of clustering algorithm randomness}

\begin{figure}
    \centering
    \captionsetup[subfigure]{labelfont={color=black}}
    \subfloat[\blue{\red{seed 0}}]{\includegraphics[width=0.35\linewidth]{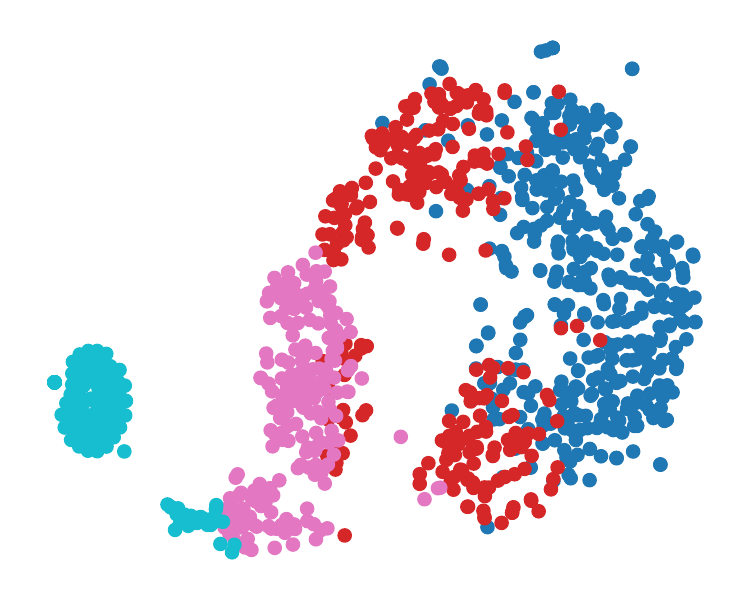}}
    \subfloat[\blue{\red{seed 1}}]{\includegraphics[width=0.35\linewidth]{pics/brats-random-0.pdf}}\\
    \subfloat[\blue{\red{seed 2}}]{\includegraphics[width=0.35\linewidth]{pics/brats-random-0.pdf}}
    \subfloat[\blue{\red{seed 3}}]{\includegraphics[width=0.35\linewidth]{pics/brats-random-0.pdf}}
    \captionsetup[subfigure]{labelfont={color=black}}
    \caption{\blue{Clustering results of SCE using different random seeds. \rred{Each color corresponds to a cluster group.} Our SCE module demonstrates robustness to the randomness of clustering algorithms.}}
    \label{fig:cluster-random}
\end{figure}

\blue{To address potential concerns regarding the influence of the clustering algorithm's inherent randomness on the SCE module's performance, we conducted experiments using multiple random seeds for clustering. The results, visualized in Fig.~\ref{fig:cluster-random}, demonstrate that for a given set of features, the clustering algorithm exhibits robustness to variations in random seed initialization. And the segmentation results also remained highly stable, with all metrics varying by less than $2\%$. Consequently, our SCE module is not sensitive to the randomness introduced by the clustering process.}

\begin{table}
  \caption{{Comparison with other refinement methods on BraTS 2019 dataset.}}
  \label{tab:other-aff}
  \centering
  \begin{tabular}{c|cccc}
  \toprule
  Refinement & \blue{DSC $\uparrow$} & \blue{Jaccard $\uparrow$} & \blue{ASSD $\downarrow$} & \blue{HD95 $\downarrow$} \\ \midrule
  PSA & $76.18$ & $72.85$ & $6.92$ & $34.40$ \\
  CRF & $73.05$ & $68.69$ & $6.80$ & $32.81$ \\
  \cellcolor[HTML]{EFEFEF}PAM (Ours) & \cellcolor[HTML]{EFEFEF}$\boldsymbol{79.69}$ & \cellcolor[HTML]{EFEFEF}$\boldsymbol{74.06}$ & \cellcolor[HTML]{EFEFEF}$\boldsymbol{5.57}$ & \cellcolor[HTML]{EFEFEF}$\boldsymbol{28.34}$ \\ \bottomrule
  \end{tabular}
\end{table}

\subsubsection{Comparison with other refinement methods}
Efforts have been made to enhance the CAM by exploring various methodologies. These methods can be broadly classified into two primary categories: those that necessitate additional training to incorporate structural information such as Pixel-level Semantic Affinity (PSA)~\cite{ahn2018learning}, and those that do not require additional training, like the Conditional Random Field (CRF)~\cite{zheng2015conditional,vemulapalli2016gaussian}, yet fail to utilize structural information. 

Contrarily, our proposed PAM module is designed to extract structural information, at the mean time without necessitating any supplementary training. 
To substantiate the effectiveness of the PAM module, we undertook a comparative analysis against the previously mentioned methods. As depicted in Tab.~\ref{tab:other-aff}, our results demonstrate that our proposed methodology outperforms the others.

\subsubsection{Analysis on small target segmentation}
\label{sec:small-target}
\begin{figure}
    \centering
    \includegraphics[width=0.9\linewidth]{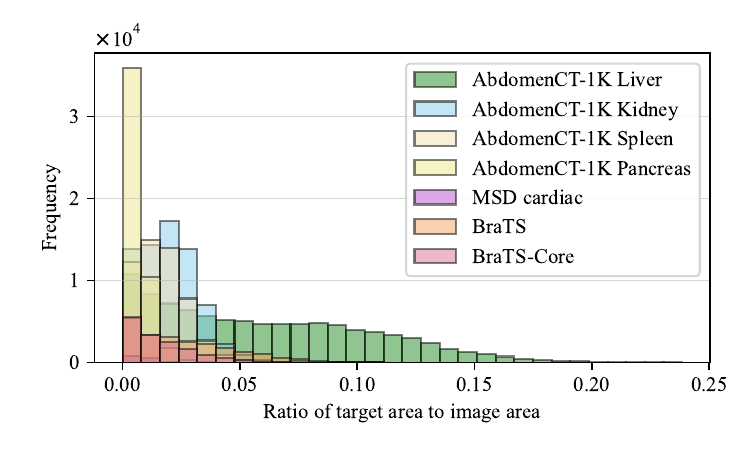}
    \caption{\blue{Ratio of target area to image area per slice for various datasets. Our experimental datasets are generally much smaller big than organs like the Liver in AbdomenCT-1K, validating our method's performance on small targets.}}
    \label{fig:area_size}
\end{figure}

\blue{
Our method's performance was evaluated across multiple datasets. And as Fig.~\ref{fig:area_size} shows, the BraTS dataset exhibits small average relative target size. Detailed experiments, including ablation studies and hyperparameter sensitivity analyses, were conducted on the BraTS dataset, demonstrating our method's effectiveness on small targets like small tumors.
}

\begin{figure}
    \centering
    \includegraphics[width=0.9\linewidth]{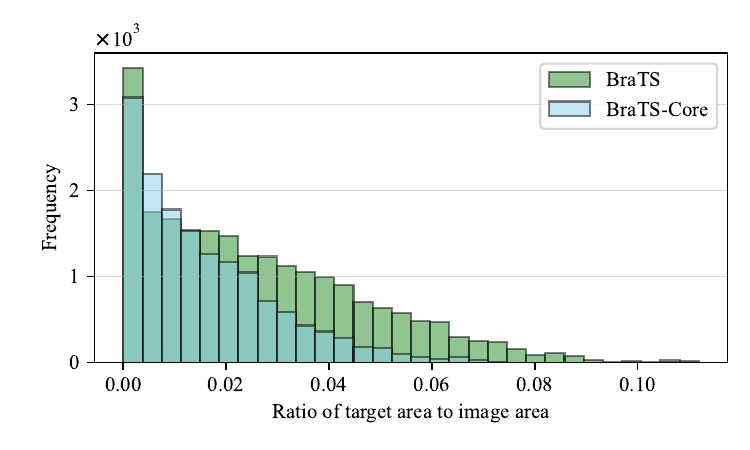}
    \caption{\blue{Ratio of target area to image area per slice for BraTS and BraTS-Core. The segmentation targets in the BraTS-Core dataset are significantly smaller than those in the already small BraTS dataset, further demonstrating the effectiveness of our method on small tumors.}}
    \label{fig:brats-vs-brats-core}
\end{figure}

\begin{figure}
    \centering
    \includegraphics[width=\linewidth]{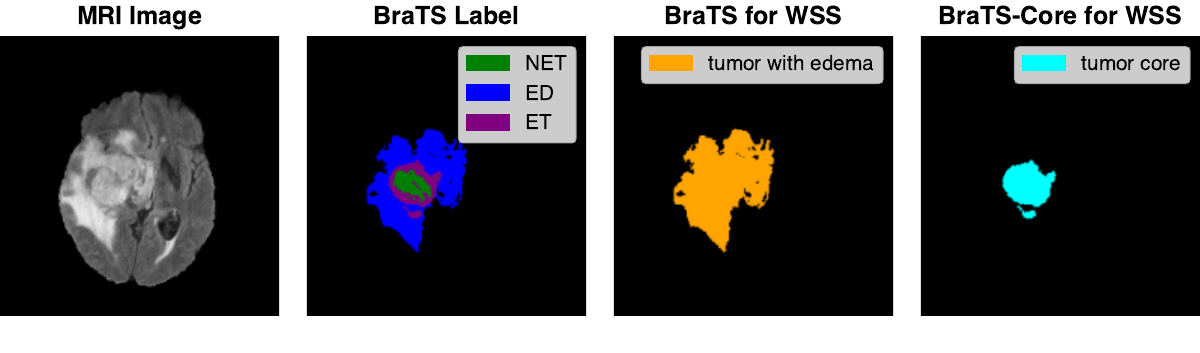}
    \caption{\blue{Comparison of the labels in the BraTS and BraTS-Core dataset. The BraTS-Core dataset exclusively utilizes the non-enhancing tumor core (NET) and the GD-enhancing tumor (ET) as labels, omitting the peritumoral edema (ED).}}
    \label{fig:brats-core}
\end{figure}

\begin{table}
    \centering
    \caption{\blue{The comparison with other WSS methods on BraTS-Core, along with the fully supervised upper bound.}}
    \label{tab:brats-core}
\blue{
    \begin{tabular}{c|c}\toprule
           Method   & DSC $\uparrow$ \\\midrule
       SIPE~\cite{chen2022self}  & 54.22  \\ 
       TOCO~\cite{ru2023token}  & 51.90  \\ 
       \cellcolor[HTML]{EFEFEF}WeakMedSAM  & \cellcolor[HTML]{EFEFEF}\textbf{61.85}  \\ \midrule
       {Upper Bound (TC)}  & {70.11}  \\ \bottomrule
    \end{tabular}
    }
\end{table}

\blue{
To further validate performance on small targets,  as shown in Fig.~\ref{fig:brats-core}, BraTS annotations were modified to retain only the non-enhancing tumor core (NET) and GD-enhancing tumor (ET), excluding peritumoral edema (ED), creating the \textit{BraTS-Core} dataset. BraTS-Core has a even smaller average relative target size than BraTS as depicted in Fig.~\ref{fig:brats-vs-brats-core}. Segmentation performance on BraTS-Core was compared with recent WSS methods and the fully-supervised upper bound in Tab.\ref{tab:brats-core}. \rred{The upper bound is defined as the mean value of the DSC metric for the tumor core (\texttt{Dice\_TC}) from the BraTS 2019 leaderboard~\footnote{\blue{https://www.cbica.upenn.edu/BraTS19/lboardValidation.html}}}. Our method shows significant advantages for small targets. However, a performance gap remains compared to fully supervised segmentation, indicating areas for future improvement.
}

\begin{table}[t]
\centering
  \setlength{\tabcolsep}{5pt}
\caption{\blue{Comparison of Encoder Parameter Sizes for Various SAM Architectures.}}
\label{tab:sam-complexity}
\blue{
\begin{tabular}{c|ccc|cccc}
\toprule
       Model  & \multicolumn{3}{c|}{SAM~\cite{kirillov2023segment}} & \multicolumn{4}{c}{SAM2~\cite{ravi2024sam}}    \\ \midrule
     Scale   & Base   & Large   & Huge  & Tiny & Small & Base & Large \\ \midrule
Size(M) & 91.0   &  308.0  & 636.0 & 38.9 & 46.0  & 80.8 & 224.4 \\ \bottomrule
\end{tabular}}
\end{table}

\subsubsection{Further numerical analysis of computational complexity}
\label{sec:numeral:complexity}
\blue{To further analysis the computational complexity of our method, we reiterate that the overhead introduced by our proposed modules is minimal. The SCE module's classification heads consist of only a few MLPs, while the PAM module leverages the prompt encoder and mask decoder from SAM, which are designed for lightweight operation, even within a web browser. The primary computational cost arises from SAM's image encoder, which remains largely frozen during our training process. Importantly, our approach can be applied as a plug-and-play enhancement to any SAM-like architecture. For a quantitative assessment, Tab.~\ref{tab:sam-complexity} provides parameter counts for commonly used SAM-like models.}

\begin{table}
    \centering
  \setlength{\tabcolsep}{5pt}
    \caption{\blue{Performance comparison of our method with the fully supervised upper bound on the BraTS dataset.}}
\blue{
    \begin{tabular}{c|cccc}\toprule
           Method   & DSC $\uparrow$ & Jaccard $\uparrow$ & ASSD $\downarrow$ & HD95 $\downarrow$  \\ \midrule
           WeakMedSAM  & \cellcolor[HTML]{EFEFEF}79.69 & \cellcolor[HTML]{EFEFEF}74.06 & \cellcolor[HTML]{EFEFEF}5.57 & \cellcolor[HTML]{EFEFEF}28.34 \\ \midrule
      {Upper Bound (WT)}  & {80.99} & - & - & {19.43} \\ \bottomrule
    \end{tabular}
    \label{tab:upper-bound}
    }
\end{table}

\subsubsection{Comparison with fully supervised upper bound}

\blue{
We compare the performance of our weakly supervised method with the fully supervised upper bound on the BraTS dataset. \rred{Being consistent} with Tab.~\ref{tab:brats-core}, we utilize the average of the DSC and HD95 metrics of this upper bound. As shown in Table~\ref{tab:upper-bound}, while fully supervised methods inherently exhibit superior segmentation performance, our approach significantly reduces the need for extensive data annotation without resulting in complete performance failure, demonstrating its potential for practical clinical applications.
}


\begin{figure}
    \begin{minipage}{0.66\linewidth} 
        \centering
        \blue{
        \includegraphics[width=\linewidth]{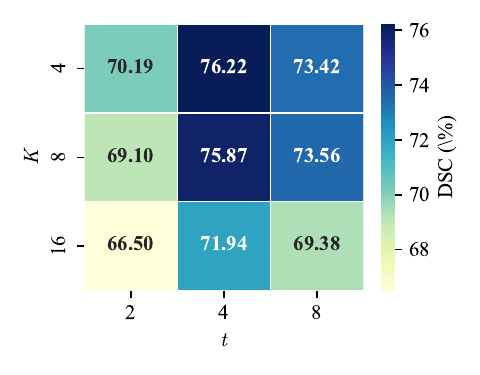}
        }
    \end{minipage}
    \begin{minipage}{0.33\linewidth} 
        \caption{\blue{\red{Evaluation of hyperparameters on the AbdomenCT-1K dataset. Results demonstrate optimal performance when $K=4$ and $t=4$.}}}
        \label{fig:abdomenct1k-param}
    \end{minipage}
\end{figure}

\subsection{Parameter Sensitivity}

\blue{To maintain a focus on the core method rather than extensive training optimization tricks, our hyperparameter ablation studies and sensitivity analyses were performed exclusively on the BraTS dataset. The hyperparameter set derived from these analyses on BraTS was subsequently applied to both the AbdomenCT-1K and MSD Cardiac datasets to evaluate the generalizability of the chosen settings.}

The method of utilizing affinity maps to obtain the probability matrix $T$ for the random walk encompasses involves hyperparameters. In Eq.~\eqref{eq:t_matrix}, the hyperparameter $\beta$ greater than $1$ is for suppressing insignificant affinities in $A$. And in Eq.~\eqref{eq:random_walk}, the hyperparameter $t$ determines the number of iterations of the random walk. An extensive set of experiments has been conducted methodically to evaluate the influence of these hyperparameters. Our empirical investigations revealed that the refinement of the affinity exhibits maximum efficacy when the threshold of random walk correlation variable $\beta$ is set to $4$, and $t$ is also set to $4$. 

\blue{As a heuristic study to determine optimal parameters for other datasets, an additional hyperparameter analysis was conducted for the AbdomenCT-1K dataset, focusing on the number of sub-classes \( K \) in the SCE module and the number of random walk iterations \( t \) in the PAM module. As a multi-organ segmentation dataset, AbdomenCT-1K has more primary classes, reducing the significance of intra-class co-occurrence compared to BraTS with only single primary class. Thus, as Fig.~\ref{fig:abdomenct1k-param} shows, a smaller \( K \) is optimal, with \( K = 4 \) yielding the best performance while \( K = 8 \) for BraTS. The optimal \( t \) remains consistent across datasets, with \( t = 4 \) performing best.}

\begin{figure}
  \centering
  \includegraphics[width=\linewidth]{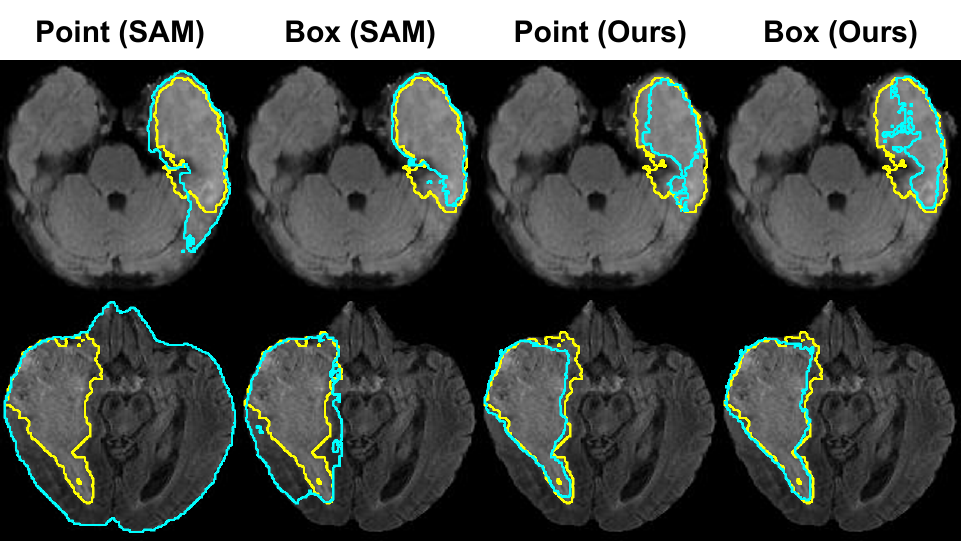}
  \caption{\blue{Comparison between SAM and interactive WeakMedSAM. The \textcolor[RGB]{220, 220, 0}{\textbf{yellow}} line represents the \textcolor[RGB]{220, 220, 0}{\textbf{ground truth}}, and the \textcolor[RGB]{0, 200, 200}{\textbf{cyan}} line indicates the \textcolor[RGB]{0, 200, 200}{\textbf{segmentation result}}.}}
  \label{fig:new_interactive}
\end{figure}
\subsection{Interactive WeakMedSAM}

Unlike fully-supervised methods such as MedSAM, which leverage massive pixel-level labels for intensive training, WeakMedSAM relies on image-level supervision, which inherently limits its interactive segmentation performance. Our objective is to demonstrate that the weakly supervised fine-tuning of WeakMedSAM does not compromise the coherence between the image encoder and the other components of the SAM architecture. To validate the applicability of WeakMedSAM in interactive segmentation scenarios, we designed two types of prompts: \emph{a single point prompt} derived from the centroid of the segmentation label and \emph{a bounding box prompt} encompassing the entire segmentation label. Comparative experiments against the original SAM, illustrated in Fig.~\ref{fig:new_interactive}, reveal that both SAM and WeakMedSAM perform well with the bounding box prompt. However, when utilizing the single point prompt, WeakMedSAM significantly outperforms SAM, demonstrating its superior capability in leveraging limited interactive information. These findings suggest that fine-tuning SAM solely using image-level classification labels to develop a model suitable for interactive segmentation is feasible; however, we acknowledge that WeakMedSAM was not specifically designed for this application, indicating a need for further research to enhance its performance in such contexts.

\section{Conclusion}

\label{conclusion}

In this study, we investigate the paradigm of weakly supervised medical image segmentation, under the guidance of Segment Anything Model (SAM). The proposed model, namely WeakMedSAM, consists of two modules: 1) the first module, \ie, SCE, involves a weakly supervised fine-tuning of SAM's ViT encoder to harness sub-class features, effectively eliminating potential co-occurrences aiming to obtain a reliable result of CAM. 2) The second module, \ie, PAM, executes a random-walk refinement of the CAM, utilizing a promptly-emerged affinity map to enhance inner-class representation. Experimental results, derived from three extensively utilized benchmark datasets, endorse the promising performance of the proposed WeakMedSAM model in this paper. In a more comprehensive context, our method also presents an innovative perspective for adapting SAM by reducing dependency on large volumes of precisely annotated data to various downstream domains, which could be adopted in other scenarios. \blue{\red{Future work will explore the potential of using a weakly supervised approach to fine-tune SAM end-to-end, potentially further improving performance and generalizability, which will focus on improving the synergy between the ViT encoder fine-tuning in the SCE module and the affinity-based refinement in the PAM module, ensuring a more cohesive and efficient adaptation of SAM to interactive medical image segmentation.}}

\bibliographystyle{IEEEtran}
\bibliography{tmi}

\end{document}